\documentclass[10pt,a4paper]{article}
\usepackage{theapa}
\usepackage[latin1]{inputenc}
\usepackage{amsmath}
\usepackage{amsfonts}
\usepackage{amssymb}
\usepackage{graphicx}

\title{Turing Test Revisited: A Framework for an Alternative}

\author{Aladdin Ayesh \\
Faculty of Technology\\
De Montfort University, Leicester LE1 9BH \\
Email: aayesh@dmu.ac.uk}

\date{July 4, 2013}

\begin{document}

\maketitle

\abstractname{:
This paper aims to question the suitability of the Turing Test, for testing machine intelligence, in the light of advances made in the last 60 years in science, medicine, and philosophy of mind. While the main concept of the test may seem sound and valid, a detailed analysis of what is required to pass the test highlights a significant flow. Once the analysis of the test is presented, a systematic approach is followed in analysing what is needed to devise a test or tests for intelligent machines. The paper presents a plausible generic framework based on categories of factors implied by subjective perception of intelligence. An evaluative discussion concludes the paper highlighting some of the unaddressed issues within this generic framework.}

\section{Introduction}

Alan Turing was one of the first provocateurs of the possibility that machines can be truly intelligent \cite{Turing1948}, \cite{Turing1950}, can think \cite{Turing1952} and can exhibit human characteristics. In his various writings, he advocated that machines could be as intelligent as a human including the abilities of creativity and emotions. As a result, he outlined a test, which became known as the Turing Test that can be applied to prove if a machine can exhibit true intelligence.

The Turing Test has a solid stand in the Artificial Intelligence research community as the ultimate test for intelligent machines. That solid stand may be the result of historical reasons. Alan Turing visionary paper and predictions make the Turing Test at the heart of any discussion on machine intelligence. Perhaps because there is no machine that can pass the test without cheating, to be controversial. One may argue that a full intelligent machine cannot pass the test but a well programmed machine within a time limit can fool a human examiner and pass the testÕs mechanism but not in its spirit. The declared aim of the Turing Test is to test intelligence unhindered by prejudice and thus to provide a benchmark by which we can tell a machine is intelligent. But does it do that?

There has been a growing body of reviews and research on the Turing Test (TT), its suitability and applicability with several attempts to augment the TT or replace it. In this paper, some of these attempts are reviewed in section \ref{review} as part of an overall but brief revisiting of the Turing Test which dominates section \ref{TT}. In section \ref{mental-physical} the issue of embodiment is addressed briefly to emphasize its importance in determining intelligence. This leads to the second major part of this paper, sections \ref{framework-1} and \ref{framework-2}, which covers the proposed framework for intelligence testing. A critical evaluation follows to conclude the paper. 

\section{Turing Test Revisited}
\label{TT}

\subsection{An Appetizer Discourse on the Turing Test}
\label{review}

In the author opinion, it is time to challenge the Turing Test, not by completing the test successfully but by questioning it suitability. In the last few years, recent advances in cognition from psychology \cite{Gross92,Solso01,Strongman2000} to computational cognition \cite{Ayesh2004b,Ayesh03,Ayesh2007,Blewitt2008,Ventura2007,Downs1977,Duric2002,Kaber2006} to mention but few, give us a new broader understanding of human intelligence, consciousness and psyche in a way that was not available to Alan Turing. In addition, the views and perspectives presented in recent years change many of the assumptions made in the past and drive us ever closer to computational models of the mind than the mere descriptive theories that were presented in the past. Experimental psychology adds to this drive towards a systematic investigation and near algorithmic explanations of the different mind processes. As a result, while no one can deny the grand impact of the Turing Test on driving the research in machine intelligence forward by providing focal point of challenge, one must question the appropriateness of the Turing Test as the ultimate test of machine intelligence. However, such questioning must not be made vainly but after a proper analysis of the essence of the Turing Test showing why such questioning may be warranted.

Similar position or thoughts were expressed in a number of papers. Some showed clear criticism to the Turing Test \cite{Krol1999,Mueller2008}, others attempt to move to a more complete or new interpretation of the Turing Test, which may lead to new tests, \cite{Hernandez-orallo99beyondthe,Alvarado2002,Dowe98anon-behavioural,Hicks2008,Bringsjord00animals}. Whilst others are firmly convinced by its validity and timelessness even though they may be somewhat critical of its interpretations \cite{Harnad92theturing,St97estimatingan,Tu97aturing}. Regardless of the viewpoint taken by many, the Turing Test has been providing means of system evaluation and inspirations to new computational techniques. One can easily observe the exponential increase of the Turing Test influence in cyber security developments \cite{Browne1991,Coates2001,ShiraliShahreza2008,Shirali-Shahreza2007d,Pope2005,Shirali-Shahreza2008} to mention but few.

\subsection{Analysis of the Turing Test}
\label{analysis}
Let us examine the Turing Test closely. For simplicity, we will choose what is often applied as the Turing Test without delving too deeply into the philosophical arguments provided in TuringÕs paper and in the rich body of philosophical discourse that surrounds some of itÕs concepts. Simply interpreted, the test requires that a human examiner to have a conversation with two unseen entities. One of these entities is a human whilst the other is the machine to be tested. There is often an agreed time limit of 5 minutes but that is often argued against. The key here is that the human judge could not tell which of the two entities is the machine and which is the human through the conversation, thus we can say indistinguishability has been pertained. Now, one of the conversation topics that we can trap a pretending machine is the weather.

Assume we asked how is the weather outside and we got some of the following for an answer:
\begin{itemize}
\item It is 21 degrees with northerly wind at speed of 5 knots
\item It is a lovely weather today [do not you like it sunny?] (the actual weather outside is 'heavy rain')
\item I do not like the weather in England, how do you cope with it?
\end{itemize}

Now which one may be a human answer and which one is a machine's? The first answer gives an impression of a machine with good weather sensors but could not be a human with a weather station who is mentally lazy and read it as it is? The last one could be a human who is foreigner to England but could not be a machine who just has a preset answers to divert a given topic to directions in which it can converse? The answer in the middle is the interesting one. The first part of the answer, which can be a genuine answer by an intelligent being be it a human or a machine, gives the impression of a machine. However, when the optional part is added, which could again be a preset answer for a machine, it gives the sense of cynicism that one likely to connect with a human rather than with pattern matching machine. These cases show the flaws in the Turing Test argument and return us to the question, what does it test?

For an ultimate intelligent machine to pass the test, the machine has to be able to pretend to be human. One may argue that this type of test requires the machine to be conscious of itself that it is a machine. It is conscious of the fact that the test requires from it to come across as a human. It is conscious of time and visual limitation. And finally it is conscious of what makes a human comes across as human, i.e. all non-intelligent human quirkiness. After all we would be much quick to accept a robot to be intelligent if it can hold a conversation with a good laugh about football! 

Given this kind of test, how many humans may fail this test and how often may they be labelled as machines, one wonders? 

\subsection{Testing Intelligence vs. Consciousness}

One claim one may make is that the Turing Test does not test intelligence, or at least not solely so. It tests consciousness, self-awareness, and the ability to lie. The last is the most important because the ability to lie is distinctively a human characteristic associated with our ability to create from imagination.

Our complex cognition makes it difficult for us to distinguish between awareness, consciousness, thinking, intelligence, and recognition of cognitive processes. The latest is a good example of the complexity and the level of interweaving of our abilities. When we remember an experience we recognize that we remembered after the memory have been recalled; but that recognition in itself make us aware of the process of memorizing; this often leads us to analyse the memory, the memorization process and the reasons why it was triggered; in other words, we become conscious of our existence in time, the existence of the memories associated with an experience and the stimulus that triggered these memories. This leaves us wonder where is intelligence in all of this and how can we quantify it for measurement?

The thoughts provoked by this article are not completely new. Similar notions of wonderment have been expressed over the Turing test and some attempts are being made to find a quantifiable test of intelligence. The advances in cognitive systems make the need to such test, or even better metrics, the greater and more urgent. Many of these alternatives, however, fail because of their focus on one element of intelligence or cognition, often focusing on learning and rational deduction. In most cases, intelligence is the result of integration of abilities, simple they may be, but together demonstrate the various facades of cognition and intelligence. For example, survival is an important ability but not necessary rational; social relations are important element of thinking but may not lead to rational decisions, e.g. parents staying with their children in a burning building.

Integrative (artificial) intelligence would require quantifiable metrics by itself measuring the different factors in ratios proportioned to their impact on behaviour. For example, learning can be form of categorization, but categorization is in itself can be form of thinking and decision making, though it may lead to stereo type based perception. Equally, categorization can be viewed as a form of memory organization to enable associative memorization. Thus, learning, thinking, memory, perception are all necessary in defining intelligence. In addition, embodiment is as important. Studies in animal intelligence gave us and could give us more insight in the separation between intelligence and the other aspects of mind and indeed of being a human.


\subsection{Defining intelligence and its typology} 
\label{typology}

Defining intelligence is a complex if not impossible task. It is not only difficult to find two people to agree on a single definition of intelligence but even one person may give multiple definitions if that person was asked the question at different times and under different circumstances. What one may provide as a definition of intelligence is often a description of the symptoms of intelligence such as:
\begin{itemize}
\item the ability to reason and solve problems (i.e. rational intelligence);
\item the ability to interact, negotiate, talk or use language intelligently (i.e. social or linguistic intelligence);
\item the ability to act in a rational way with consideration under given circumstances or stimulus (i.e. behavioural intelligence);
\item the ability to learn, adapt, create new ideas, and innovate (i.e. constructive intelligence)
\end{itemize}

Interestingly enough, one would rarely, if ever, receive emotions, love, ability to marry and reproduce, and other humane activities as features or definitions of intelligence whilst they are at the core of distinguishing human and living entities intelligence. One may go as far as declaring these humane features are the essence of the most controversial and contested concept-notion 'the Soul'.

One important outcome of the list above is the different types of intelligence one may find in describing this elusive notion. This typology of intelligence is most important in designing intelligence tests for humans and machines. If one may hope to test true intelligence one should aspire to test its different facades. One other notion should be mentioned here before discussing the concept of true intelligence is that levels of intelligence demonstrated. To give an example, consider the intelligence of an autistic child, whilst the child may demonstrate high level of intelligence in some types mentioned that child may have diminished intelligence levels in others. Similarly, an argument can be made about conscious and unconscious levels of intelligence. This subject requires a discourse of its own to be had at another time.

\subsection{Is true intelligence useful?} 
\label{true}

For the sake of completeness, a discourse on the truthfulness of intelligence is needed. If one is asked the question labelling this section: 'is true intelligence useful?' one may answer hastily with yes; but such an answer needs a careful consideration and reflection. To present the essence of the question on truthfulness more appropriately, let us ask: is Terminator had true intelligence? Does a human warrior have a true intelligence? The answer has to be somewhat is yes and yet we are not sure if that true intelligence is useful. Usefulness and trueness of intelligence are two different things. What one may perceive as true intelligence is because a human performed the tasks associated with that intelligence but equally that so perceived true intelligence may have far disastrous consequences to be comprehended or explained rationally. In fact that kind of intelligence is distinguished in relation to human actions from what is commonly meant by the word intelligence by labelling the former with terms such as 'stupid', 'inexperienced', 'incompetent', and so forth. Interestingly, the abstract form of intelligence is never denied as a feature associated with the actor of such stupid actions. It seems when humans are concerned there is a distinction made between rationality and intellect on one hand and day-to-day intelligence on the other. Nonetheless, this may not answer the initial question of 'is true intelligence useful?'.

The relationship between the truthfulness of intelligence and its usefulness is somewhat a false subject of discourse. Let me explain. Let one assumes human intelligence to be the true form of intelligence as one may define intelligence. Now, part of that intelligence is emotions, social relations, and their impacts demonstrated by behaviours of no relevance to rationality. To give an example, is spending 100 considered overspending? What about 100 million? Regardless of the answer given, the reality is that it is all relative. If someone earns a billion pound a year, what is a 100 million spending? But a 100 pound out of the pocket of someone earning 500 pounds a month is a huge amount of money. The aim of this example is to show that pure logical rationality does not play a factor in every human decision but there are many life factors that play a role. The ability to explain one behaviour such is the case in the example does not qualify the behaviour to be rational and logical in the strict sense of the words. Nevertheless, none of the above denies human intelligence. What may be perceived as intelligence is the result of the interaction of this multitude of factors governed by emotional rules. These emotional rules in interaction with experiences presented in the form of beliefs and perceptions provide the reasoning engine behind human intelligence, thus behind true intelligence. Whilst that true intelligence not necessary be useful in itself, in fact one may argue it is usefulness neutral, this irrational intelligence is the core of all useful actions human agent may take and perform.

\begin{figure}[htbp]
   \centering
   \includegraphics[scale=0.3]{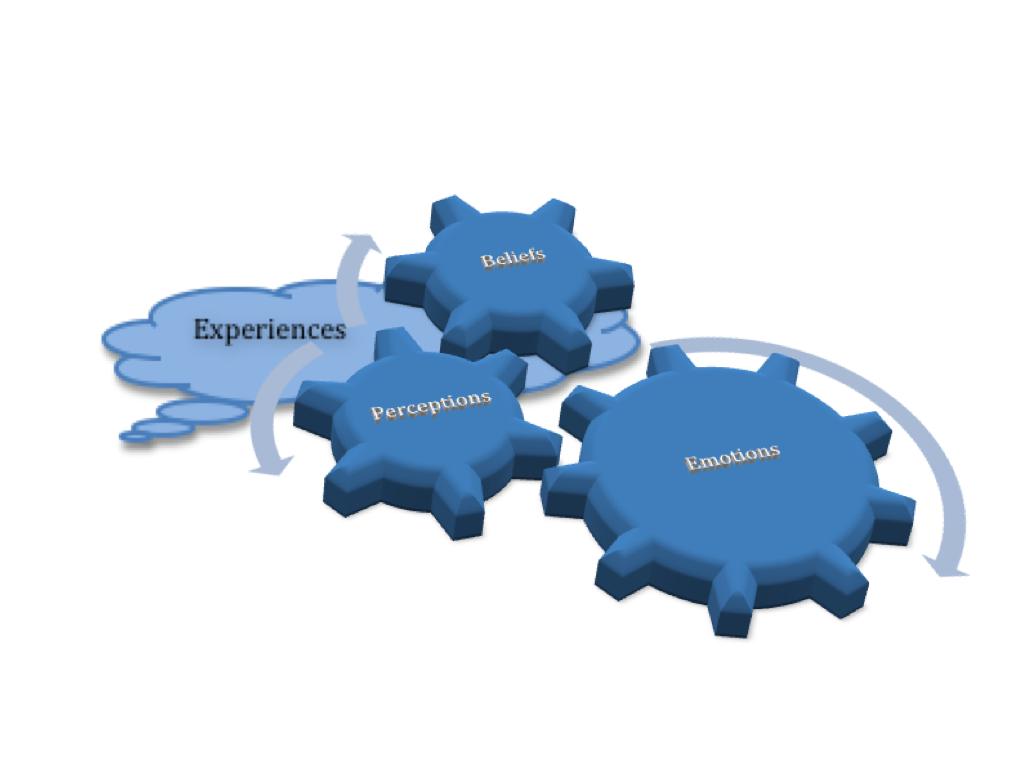} 
   \caption{Human Intelligence Engine of Irrationality}
   \label{mindEngine}
\end{figure}

Figure \ref{mindEngine} shows the human intelligence engine. Beliefs and perceptions are influenced consciously or unconsciously by experiences being physical experiences, e.g. getting bitten by a dog as a child makes one afraid of dogs, or by verbal inherited experiences, e.g. cultural norms of not eating insects in the west and cockroaches being filthy. While the three notions are often associated with irrationality, one can argue strongly with a long list of examples these are the bedrocks of humans decision-making, memorization and even cognition.

\section{Mental vs. Physical states}
\label{mental-physical}

In relating the mental image of an agent's state and its physical existence, one can do far worse than reading Longo's paper on life phenomena \cite{Longo2009}. His bio-physical approach to the question of organisms and programs, structure and circularities, gives a good starting point to the need of distinguishing these two level of conscious representation: mental and physical. Yet, one cannot take one without the other, which for years seems what is happening in artificial intelligence community and research. Perhaps one should emphasis the term of Integrative Intelligence as what artificial intelligence research should be aiming to achieve. However, this is not the place for such argument. 

With a reference to the discussion to come in section \ref{factors}, one may assert that no true intelligence can be found without a physical embodiment through which this intelligence is expressed in a behaviourist manner. The behaviour could be positive type where a selected, rationally or otherwise, and executed action takes a place; or it could be a passive where the embodiment formation indicates the behaviour and express the thought, e.g. involuntary facial expression or body gesture.

It is clear that physical states are driven somewhat from mental states. The author opinion is that the difference between a rational selection of an action and a reactive or passive behaviour is the difference between conscious and unconscious translation of the physical representation of the mental state of an intelligent agent. 


This relation between consciousness and the representation of different states of being can be reduced to self-diagnosis system. Within such system different levels of self-awareness is represented through a state, e.g. mental state, emotional state, physical state, spatial state, and so forth. Each state can be monitored. The interaction between these different state is of great importance and the difficulty of realizing true intelligence relies in the complexity of representing and implementing such intertwined relations between these different awareness states. One of the few attempts to do so can be seen in the primitive hierarchical cognitive map representation used in relating emotions to physical and spatial states \cite{Ayesh2004b,Ayesh03}.   

\section{Towards Computational Cognition Verification}
\label{framework-1}

\subsection{Cognitive Systems Engineering}
The increase in cognitive systems development, some of which fuelled and supported by advances in game technologies \cite{Blewitt2009}, whilst others benefited from advances in computational powers became available, requires engineering disciplines to be applied in developing and testing these systems. The birth of cognitive systems engineering has been and yet the tools and means are still in development. Agent-oriented software engineering may be seen the first signs of engineering approach to cognitive systems analysis and design but it still lacks comprehensive verification and validation tools \cite{Laufmann1997,Wooldridge1997}. In fact, one may argue agents-based systems, let alone cognition, lacks formal means of verification unless these are logic-based agents \cite{Ventura2007,Zheng2005}. 

The difficulty of cognition with formal classical logic is the long standing problem of human reasoning does not fully fit with a formal strict first order logic. The simulation of human cognition leads by its nature to conflict, inconsistency and nondeterministic of varying degrees and levels. Cognition may be compared to chaotic systems than well natured metric compliance systems. Does this mean a verification process for cognitive systems is impossible? Many aspects of cognition are observable, thus testing cognitive systems will require a balance between subjective and objective testing. Formalizing that balance in metrical form, i.e. what to be observed and expected, is needed instead of attempting to adapt formal methods techniques in verification and validation. One may say that is exactly what Turing Test is about and the response would be of agreement. However, the Turing Test fails in identifying these subjective and objective observable measures explicitly.



\subsection{The Human Factor}
The Turing Test highlighted one important factor in intelligent systems testing that is Human Factor. If we as humans accept that a system to be intelligent then it must be so. How this acceptance may be gained? According to the Turing Test is by the indistinguishability testing. One may re-phrase that to be by deceiving group of humans in believing a machine is a human. If the principle of 'you can deceive some people some of the time but not all people all the time' holds, then the success of this test is short lived. Instead, one should seek the human acceptance of an intelligent machine as that an intelligent machine. This can be achieved by making humans project and relate to the machine being tested.

In a work done using AIBO robots to stimulate creative process during a writing workshop, conducted under AIBOStories title, the shape and behaviour of the robot had influenced in the participants a 180 degree turn from cynicism to positive interaction. This effect came about as a result of the shape of the robot, i.e. projection, the behaviour and responses of the robot, i.e. expression and emotions, and the interaction capabilities although simple was still enough to feed into the other aspects, i.e. social interaction and integration. Regardless of the complexity of the internal mechanisms, these observable factors influenced the reaction towards the robot, hence the acceptance of that robot as a possible pet.

\section{A Framework for Intelligence Testing}
\label{framework-2}

\subsection{What are we testing?}
It is important to be clear on what will be tested. As the discussion presented above, while the main goal of Turing Test was testing true intelligence, the test specified and its interpretations may have tested the wrong side of the equation. It often played on known weaknesses of the machine for lack of development, as for example the case with CAPTCHAs \cite{Landwehr2008,Kolupaev2008,Pope2005}.

Similarly, traditional software engineering and formal methods verification techniques fail to capture the essence of cognitive systems that makes these systems distinctive, and hence fail in testing intelligence. The discussion led to the need of metrics of subjective nature or at least a balance between objectively measurable variables and subjectively observable variables in forming a true intelligence test.

Observations and experiments \cite{Ayesh2007a} showed us that different factors play a role in accepting a system as an intelligent system from the physical form to the response form. Light, waging a ear, sound, and so forth are all elements that play a role in giving the impression of life and intelligence. One may put a proposition that \textit{wherever there is a perceived life there is a perceived intelligence}.

\subsection{Factors as Metrics} \label{factors}
Whilst there are a huge literature on the Turing Test, its many reincarnations, and attempts of analysis, few takes systematic study of the factors of true intelligence to be considered. One suspects that a look at psychological studies would be of tremendous help. Nonetheless, Alvarado and his colleagues seem to summarize some of these factors nicely in this paper on performance metrics \cite{Alvarado2002}. If one may argue with the table they present on metrics for the simulation of mind, and indeed their core argument, the criticism would be directed towards their assumption that mind can exist in abstraction without embodiment. Considering their clear focus on the social factors of intelligence, one may be surprised at this abstraction of mind.

The second point of criticism would be for their focus on associative learning and neglecting other, more subtle, types of learning that is the result of experience and reasoning. While their description of associative learning attempts to capture elements of experience and their association to behaviors, the whole section neglects the impact of emotions on memory processes, to give an example, and how emotional experiences may be recorded. While associative learning associates the experiences and the responses, how learning facts and deriving conclusions as an internal reasoning process can be counted for?

\begin{figure}[htbp]
   \centering
   \includegraphics[scale=0.75]{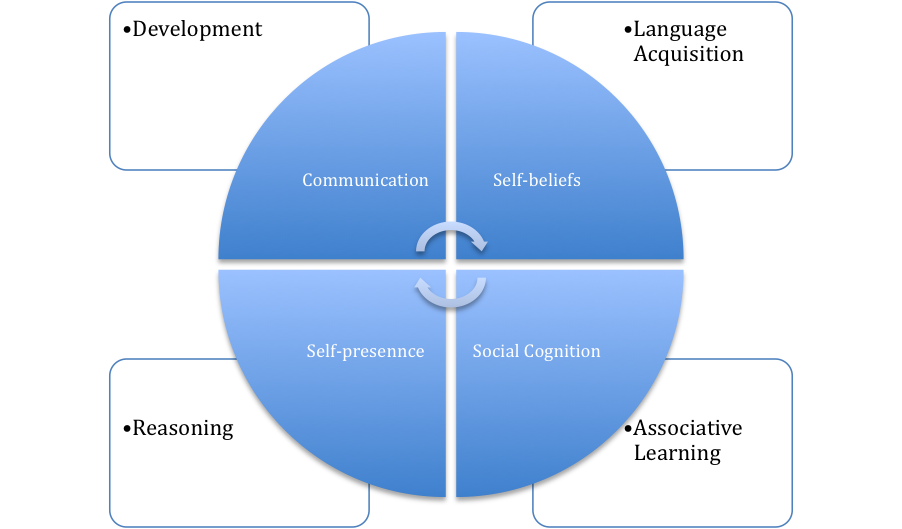} 
   \caption{Attempt to Capture the Factors of Intelligence}
   \label{cycleMind}
\end{figure}

Another aspect of learning that is more relevant to behaviors is the development of skills from elementary abilities. Whilst Alvarado et. al. account for conditioning in their metrics conscious acquisition of skills is not explicitly present. If one may look at the Social Cognition part of their table, one can identify communication and interaction as a missing part whilst representation of self and others has surprisingly a strong presence. Still this is a very interesting attempt. Figure \ref{cycleMind} attempts to build on Alvarado's table by reducing the details of some of the factors while expanding other dimensions of these factors. 

In figure \ref{cycleMind} self-presence captures the embodiment question. This includes both spatial and mental imaging of self. The extreme capability is to be able to recognize one's self in the mirror which some may take for consciousness. Self-beliefs on the other hand expand on social encoding and representation of self that proposed by Alvarado's table. This includes all aspects of belief an agent may hold about itself, others, and environment. This also encompasses emotions and perceptions where perceptions can be defined as learned associations between factual beliefs and emotional experiences in relation to the subject of a given perception, some initial discussions on this can be found in \cite{Ayesh03}.

\subsection{The 4 E's Framework}
\begin{figure}[htbp]
   \centering
   \includegraphics[scale=0.75]{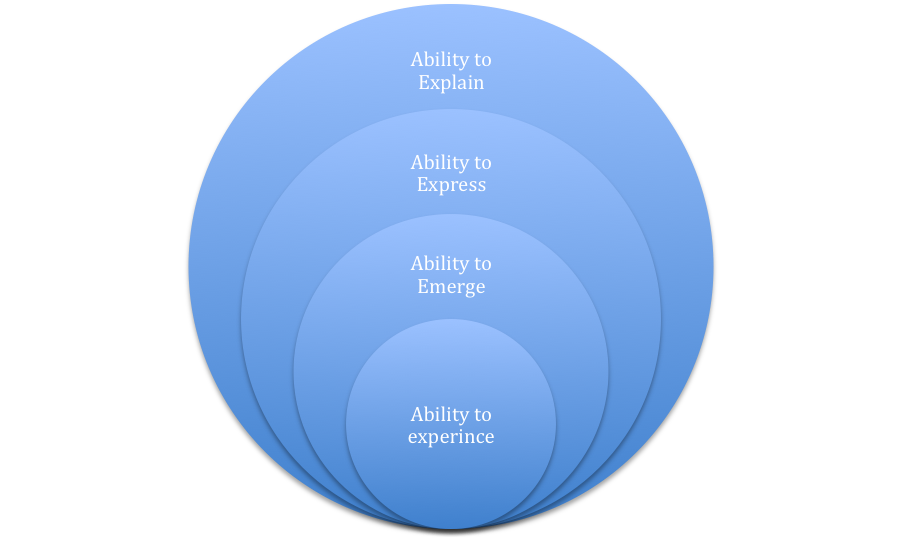} 
   \caption{The 4 E's Framework}
   \label{fourEs}
\end{figure}

The discussion presented so far in this paper leads to the specification of a new test based on factors used in determining the presence of intelligence rather than testing the system directly. Whilst \cite{Alvarado2002} provides an interesting summary of some of these factors, attempting to include all factors in developing a performance metrics will inevitably leaves some of these factors unaccounted for. In addition, it will be a ridged system for testing cognitive systems that neglect the typology of intelligence discussed in section \ref{typology}. Instead a generic framework is presented here based on classification of the factors discussed previously. For example, the ability to express will include the different levels of expression from a sound in dog-like robot to a full elaborate natural language verbal communication in a literary chatterbox. Figure \ref{fourEs} shows the four generic classifications of cognition factors.

The hierarchy presented in figure \ref{fourEs} is important and it enables the consideration of intelligence typology. The simplest of intelligence presence will be apparent through the ability of experience. Animals and human alike can experience pain, pleasure, and other sensory feelings and responses to them. That response if coupled by social rules and interactions will lead to emerging behaviour, organization and internal-external representations. As this emergence develops to more sophisticated levels the ability to express feelings, emotions, internal beliefs and so forth becomes very important and of tremendous impact on the level of intelligence demonstrated. This can be qualified by the observed results from experimenting with AIBO dogs. Their ability to express agreement and disagreement using coloured lights and movement of ears and tail had a great impact on how participants related to the robot and changed their attitudes. Finally, the ability to explain is at the ultimate level of human intelligence and in its sophisticated form one may argue it crossed the boundaries to the question of consciousness. 


\section{Evaluating the Framework}
This is a short note on evaluating the framework. By no means this is a full evaluation but more of a reflection. First of all, the framework should be possible to be instantiated to create different types of tests for different types of intelligence. The framework proposed here lacks the instantiation rules and mechanisms to ensure the correct instantiation and application. One important issue here is the minimum required testing to prove a demonstration and perception of intelligence.

Another issue is embodiment issue. It is addressed within this framework implicitly through experience and expression. However, the aesthetics of embodiment is important in itself to be addressed implicitly. An explicit examination of physical representation and expression is necessary. 

Finally, the ability of explaining could leads to imagination and eventually true machine creativity. Reaching those levels of exclusively human intelligence could outstretch the framework beyond its capabilities. 

\section{Conclusion}
In this paper, the Turing Test has been discussed and analysed in the light of literature and the author personal views. It followed from that discussion a discussion on testing intelligence and verification of cognitive systems. It became clear that an objective test of cognitive systems would be invalid test and more subjective testing is required. In studying the subjective factors one can specify certain factors impacting on perception of intelligence and the acceptance of it as being a demonstrated intelligence. Drawing again on research from literature and from the authors own work some of these factors can be listed and furthermore categorized. At the same time, the discussion led to typifying intelligence.

Intelligence typology indicates strongly toward different type of intelligence but also at different levels of intelligence. That discussion also led to the importance of integrated intelligence hence the different types of intelligence are often present together in different portions to create the mixture that is the intelligent creature or agent to be examined. As a result, relying on a listing of factors or performance metrics may not be sufficient to test different types of intelligence in its different forms and mixtures.

In conclusion, a new framework based on generic categorization of types and factors of intelligence was presented. This proposed framework could enable the instantiation of different tests that are suitable for different systems. The framework is still in development and further work on operational mechanisms of instantiation and minimum requirements is still needed.

\bibliographystyle{theapa}
\bibliography{TTest.bib}

\end{document}